\documentclass{ieeeoj}

\usepackage{cite}
\usepackage{amsmath,amssymb,amsfonts}
\usepackage{graphicx,color}
\usepackage{textcomp}
\usepackage{booktabs}
\usepackage{array}
\usepackage{siunitx}
\usepackage{tikz}
\usetikzlibrary{calc}
\usepackage{multirow}
\usepackage{makecell}
\usepackage{algorithm}
\usepackage{algpseudocode}
\usepackage{dirtree}
\usepackage{stfloats}
\usepackage[hidelinks]{hyperref}

\def\BibTeX{{\rm B\kern-.05em{\sc i\kern-.025em b}\kern-.08em
    T\kern-.1667em\lower.7ex\hbox{E}\kern-.125emX}}
\AtBeginDocument{\definecolor{ojcolor}{cmyk}{0.93,0.59,0.15,0.02}}
\def\OJlogo{\vspace{-14pt}\includegraphics[height=28pt]{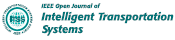}}

\begin{document}

\receiveddate{XX Month, XXXX}
\reviseddate{XX Month, XXXX}
\accepteddate{XX Month, XXXX}
\publisheddate{XX Month, XXXX}
\currentdate{XX Month, XXXX}
\doiinfo{OJITS.2026.XXXXXXX}

\title{Automated Digital Twin Construction for Highway Scenarios Using LiDAR Point Clouds and OpenStreetMap}

\author{Yongqi Zhao\affilmark{1}, Dong Bi\affilmark{1, 2}, Paul Kovacevic\affilmark{1},
Tomislav Mihalj\affilmark{1}, Martin Schabauer\affilmark{1}, Johannes Betz\affilmark{3}, and Arno Eichberger\affilmark{1},
Member, IEEE}

\affil{Institute of Automotive Engineering, Graz University of Technology, 8010 Graz, Austria (email: yongqi.zhao@tugraz.at; dong.bi@tugraz.at; paul.kovacevic@tugraz.at; tomislav.mihalj@tugraz.at; martin.schabauer@tugraz.at; arno.eichberger@tugraz.at)}
\affil{School of Intelligent Connected Vehicle, Hubei University of Automotive Technology, 442002 Shiyan, China (email: dong.bi@tugraz.at)}
\affil{Professorship of Autonomous Vehicle Systems, Technical University of Munich, 85748 Garching, Germany (email: johannes.betz@tum.de)}
\corresp{CORRESPONDING AUTHOR: Dong Bi (e-mail: dong.bi@tugraz.at).}
\authornote{This work has been submitted to the IEEE for possible publication. Copyright may be transferred without notice, after which this version may no longer be accessible.}
\markboth{Automated Digital Twin Construction for Highway Scenarios Using LiDAR Point Clouds and OpenStreetMap}
{Zhao \textit{et al.}}

\begin{abstract}
Accurate road environment modeling is fundamental to the simulation and validation of automated driving systems. However, constructing road maps in standardized formats such as ASAM OpenDRIVE from real-world sensor data remains a time-consuming and costly process. Mobile mapping LiDAR captures accurate lane-level geometry but is confined to the driven corridor, while OpenStreetMap (OSM) provides broad road network topology but lacks geometric precision at the lane level. To address this, an automated workflow is proposed to fuse LiDAR point clouds with OSM data to generate georeferenced ASAM OpenDRIVE maps of highway environments, requiring minimal manual intervention. The pipeline reconstructs mainline roads from LiDAR-derived measurements and infers ramp geometry and topology from the OSM road graph, enabling complete highway interchange modeling without full sensor coverage. Experiments demonstrate a mean lateral RMSE of 0.740 m, and the generated maps are directly usable in mainstream simulation platforms including IPG CarMaker and Esmini. These results validate the effectiveness of combining measurement-derived geometry with map-derived topology for automated OpenDRIVE digital twin generation. The project code is available at \url{https://github.com/ftgTUGraz/opendrive-digital-twin-generator}.
\end{abstract}


\maketitle

\section{INTRODUCTION}
\IEEEPARstart{S}{imulation}-based testing has become an essential tool for the development and verification of automated driving systems, especially when real-world testing is costly, time consuming, or difficult to reproduce~\cite{KALRA2016182}. A key prerequisite for such testing is a digital representation of the road environment that can support visualization, traffic simulation, and vehicle dynamics~\cite{10242366}. For automated driving, this requirement is more restrictive: the road model must not only be visually plausible, but also machine readable, containing lane geometry, elevation, and connectivity information that can be interpreted by simulation tools.


High-definition (HD) maps and structured road models provide such machine-readable representations for automated driving applications~\cite{Elghazaly2023HDMaps,Liu2020HDMap}. However, producing and maintaining these maps remains expensive because mapping pipelines typically involve dedicated measurement platforms, complex postprocessing, and manual correction. Simulation-oriented applications require road models with standardized geometry, lane structure, elevation, and topology that can be interpreted by simulation tools~\cite{ASAM_OpenDRIVE_190}. ASAM OpenDRIVE provides an open, standardized road description format widely adopted in simulation environments~\cite{11361285}. However, generating valid OpenDRIVE maps from real-world road data remains a practical bottleneck, particularly in highway scenarios involving mainline-ramp connections.


Mobile mapping LiDAR approaches have made significant progress toward automatic road model generation. Prior studies have generated OpenDRIVE road models from mobile mapping measurements by extracting lane markings from point clouds and fitting them into OpenDRIVE-compatible representations~\cite{Chiang2022OpenDRIVE, barsi2021building}. Other work has demonstrated that LiDAR point clouds can support road boundary extraction, centerline recovery, elevation estimation, and digital modeling of road scenarios~\cite{Wang2023LiDARRoadModeling}. More recent AI-assisted workflows have further reduced manual effort in HD map generation from mobile mapping data~\cite{Somogyi2025AIHDMap}. These methods demonstrate that measurement-derived data can provide accurate local road geometry. Nevertheless, these methods remain constrained by the measurement trajectory: road elements outside the measurement corridor, such as highway ramps, cannot be recovered from LiDAR measurements alone. Moreover, lane connectivity and road topology between mainline and ramp segments are omitted in existing pipelines, limiting the completeness of the generated maps in highway scenarios.


\begin{figure*}[ht]
    \centering
    \includegraphics[width=\textwidth]{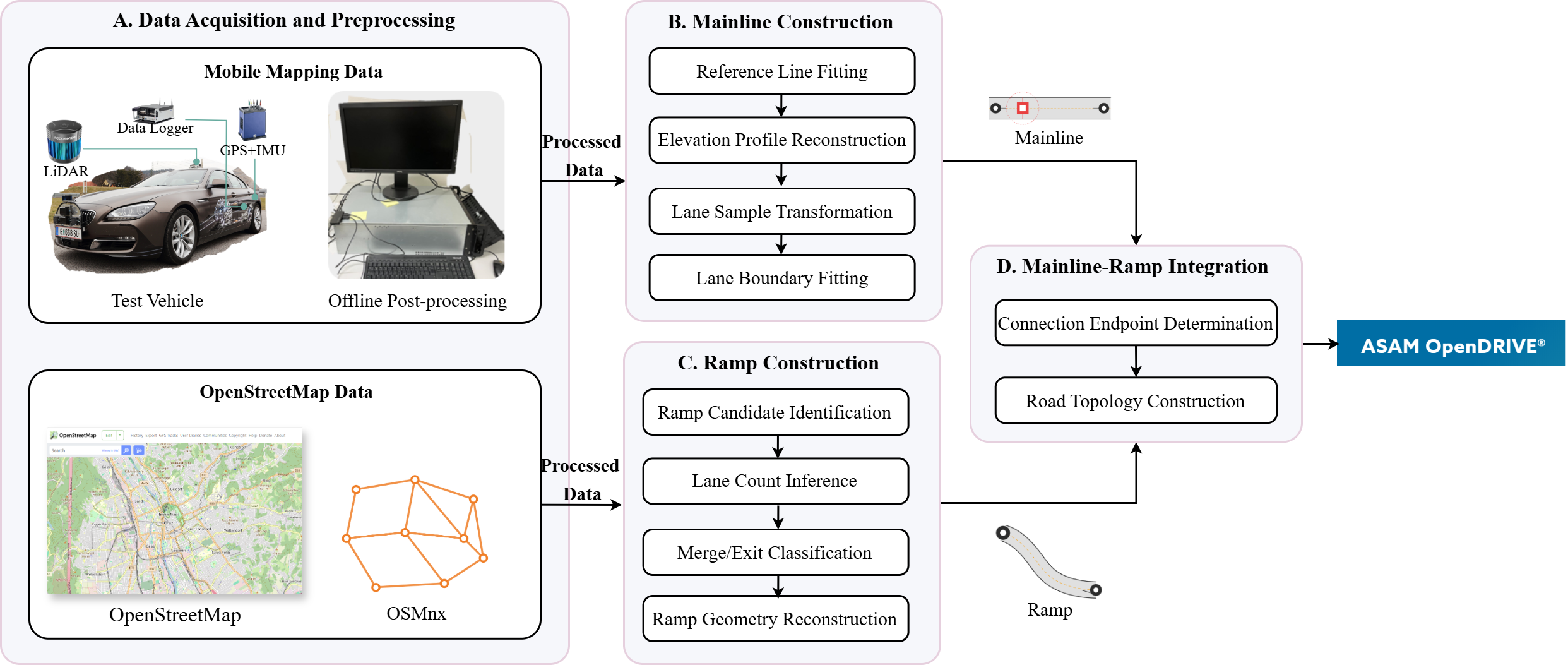}
    \caption{Overview of the proposed hybrid digital twin reconstruction workflow. LiDAR measurements are used to reconstruct the driven highway mainline, while OSM supplies the unobserved ramp geometry and topology. The two sources are integrated into a georeferenced ASAM OpenDRIVE map for simulation.}
    \label{fig:system_overview}
\end{figure*}

OpenStreetMap (OSM) provides a complementary source of road information. It contains large-scale road graph data, including edge geometries, road classes, directionality, and network connectivity. However, OSM data are not designed as lane-level simulation maps. Their geometries, lane attributes, and topology are useful as priors, but they are not sufficiently reliable to be directly used as final OpenDRIVE maps. This motivates a hybrid strategy in which mobile mapping data provide accurate local geometry for the measured mainline, while OSM provides geometric and topological priors for road elements outside the measurement corridor.


In this paper, an automated workflow is proposed for generating simulation-oriented OpenDRIVE digital twins in highway scenarios by combining mobile mapping LiDAR measurements and OSM road graphs. The main contributions of this work are summarized as follows:
\begin{itemize}
    \item A hybrid LiDAR-OSM workflow is proposed for automated OpenDRIVE digital twin generation in highway scenarios.
    \item A mainline-ramp reconstruction and integration method is proposed to combine LiDAR-derived mainline geometry with OSM-derived ramp topology and geometry.
\end{itemize}

\section{METHODOLOGY}
The proposed workflow reconstructs a highway digital twin by combining two complementary data sources, as shown in~\autoref{fig:system_overview}. Vehicle mounted LiDAR measurements provide accurate road edge, lane sample, and trajectory information for the driven mainline, whereas OSM supplies geometric and topological priors for untraversed ramp segments. The pipeline converts the LiDAR data into a mainline representation for OpenDRIVE, extracts and classifies ramp candidates from the OSM road graph, and then integrates both sources through directed topology construction and OpenDRIVE export. The final output is a georeferenced ASAM OpenDRIVE map for downstream simulation.

\subsection{Data Acquisition and Preprocessing}

\subsubsection{Mobile Mapping Data}
\label{subsec:test_vehicle}
\paragraph{Test Vehicle} The mobile mapping data were collected using a customized BMW sedan equipped with a rooftop 128-beam LiDAR, a GENESYS Automotive Dynamic Motion Analyzer (ADMA)\footnote{https://genesys-offenburg.de/en/adma-g/}, and a dSPACE AUTERA\footnote{https://www.dspace.com/en/pub/home/products/hw/autera.cfm} data logger. The LiDAR provides high density, \SI{360}{\degree} observations of the driven mainline, from which road edge point clouds and lane marking samples are extracted. The ADMA records synchronized vehicle position, velocity, and altitude, which are later used to georeference the observations from LiDAR and transform lane samples from the ego vehicle frame to the mapping frame. The data logger synchronizes and stores all sensor streams during the measurement run. The main LiDAR specifications are summarized in~\autoref{tab:lidar_params}.

\begin{table}[ht]
\centering
\caption{LiDAR's technical data~\cite{10417138}.}
\label{tab:lidar_params}
\begin{tabular}{ll} \toprule[2pt] 
\textbf{Parameter} & \textbf{Value} \\ \midrule[1pt]
Beam & 128 \\
Operating frequency & \SI{905}{\nano\meter} \\
Height (above the ground) & \SI{1.8}{\meter} \\
Transmitting power & \SI{90}{\watt} \\
Horizontal FOV & \SI{360}{\degree} \\
Vertical FOV & \SIrange{-25}{15}{\degree} \\
Beam width (horizontal) & \SI{1.5}{\milli\radian} \\
Beam width (vertical) & \SI{3.6}{\milli\radian} \\
Detection range & \SIrange{0.4}{250}{\meter} \\
Detection accuracy & \SI{3}{\centi\meter} \\ \bottomrule[2pt]
\end{tabular}
\end{table}

\paragraph{Offline Postprocessing}
\label{subsec:postprocessing}
After each measurement run, the synchronized sensor streams were transferred to an offline postprocessing server. The server processes the raw LiDAR point clouds together with the recorded vehicle position and orientation to extract structured road geometry information, including road edge point clouds, lane marking samples, per frame mapping position and orientation, and the GPS origin used for coordinate initialization. The processing was executed on a dedicated 4U rack-mounted server equipped with an AMD EPYC 7F75 processor, 128 GB DDR4 RAM, an NVIDIA RTX 3090 GPU (24 GB VRAM), and 8 TB SSD storage. These outputs provide the inputs from LiDAR for the subsequent mainline reconstruction stage.

\subsubsection{OSM Data}
\label{subsec:osm_data}
\begin{figure}[ht]
    \centering
    \includegraphics[width=\linewidth]{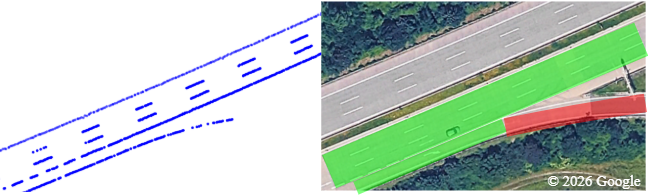}
    \caption{Illustration of LiDAR coverage limitation. Left: LiDAR point cloud acquired along the highway mainline. Right: aerial view of the study site, with the mainline highlighted in green (measured) and the ramp highlighted in red (unmeasured).}
    \label{fig:LiDAR_ramp_overview}
\end{figure}

As illustrated in~\autoref{fig:LiDAR_ramp_overview}, the mobile mapping vehicle only observes the driven mainline, and ramp segments that are not traversed during the measurement run remain unavailable in the LiDAR point cloud. OSM\footnote{https://www.openstreetmap.org} is therefore used as a complementary map source for these unmeasured road elements. In the proposed workflow, the Python package \textit{OSMnx}~\cite{boeing2025modeling} is used to query a local drivable road graph around the measured trajectory. The queried graph contains directed road edges, node connectivity, polyline geometries, and metadata at the road level.

For later ramp reconstruction, the OSM data are preprocessed to retain both geometric and semantic information. The edge geometries are projected into the same local metric coordinate frame as the trajectory from LiDAR, while metadata fields such as \textit{highway}, \textit{lanes}, \textit{lanes:forward}, \textit{turn:lanes}, and \textit{oneway} are preserved. In particular, road classes carrying the \textit{\_link} suffix provide ramp candidates, lane tags support lane count estimation, and directed graph connectivity provides the basis for merge and exit classification in the subsequent ramp reconstruction stage.

\subsubsection{Processed Data}
After mobile mapping postprocessing and OSM graph preprocessing, the processed data are organized into the items summarized in~\autoref{tab:pipeline_inputs}. These data provide the geometric, semantic, and coordinate information required for the subsequent reconstruction steps.

\begin{table*}[!t]
\centering
\caption{Data items prepared for digital twin reconstruction.}
\label{tab:pipeline_inputs}
\begin{tabular}{l l >{\linespread{0.85}\selectfont}m{0.58\textwidth}} \toprule[2pt] 
\textbf{Data Source} & \textbf{Data Item} & \textbf{Description} \\ \midrule[1pt]
\multirow{4}{*}{\centering Mobile Mapping} & Vehicle pose trajectory & Per frame vehicle position and orientation recorded during mapping. \\
& Road edge point clouds & Extracted boundary points describing the driven mainline edges. \\
& Lane sample points & Sparse lane marking observations detected in each LiDAR frame. \\
& GPS origin coordinates & Geographic origin used to define the local metric coordinate frame. \\
\midrule[1pt]
\multirow{4}{*}{\centering OSM} & Directed road graph & Local drivable road network represented by directed nodes and edges. \\
& Edge geometries & Polyline geometries describing road and ramp centerlines. \\
& Road metadata & OSM tags describing road class, lane information, and driving direction. \\
& Node connectivity & Graph connectivity describing how road edges are linked. \\ \bottomrule[2pt]
\end{tabular}
\end{table*}

\subsection{Mainline Reconstruction}
\begin{figure}[ht]
    \centering
    \includegraphics[width=\linewidth]{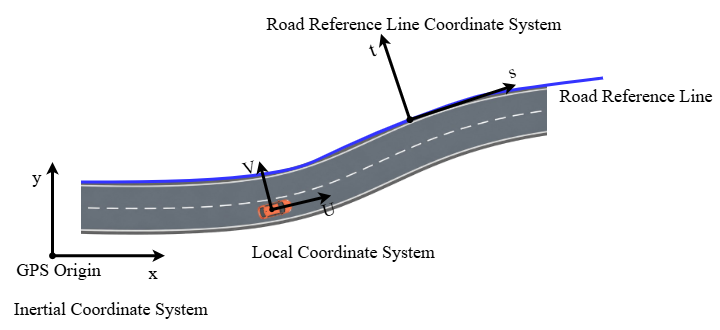}
    \caption{Coordinate frames used in the proposed workflow based on~\cite{ASAM_OpenDRIVE_190}.}
    \label{fig:coordinate_frames}
\end{figure}

This section describes how mobile mapping data are converted into a representation of the highway mainline that is compatible with OpenDRIVE. The coordinate frames used for transforming lane samples and fitting lane boundaries are illustrated in~\autoref{fig:coordinate_frames}. Road edge point clouds are used to fit the reference line and reconstruct the vertical profile, while lane marking samples are used to estimate lane boundaries. The reconstruction includes reference line fitting, elevation profile reconstruction, transformation of lane samples, and lane boundary fitting in a Frenet frame.

\subsubsection{Reference Line Fitting}
The mainline from LiDAR is reconstructed segment by segment along the measured trajectory. In the implementation, the trajectory is divided into approximately \SI{100}{\meter} windows, and only the road edge points located near the current trajectory window are selected for fitting. This local filtering avoids using unrelated point cloud regions when the road is curved or when adjacent roads are close to the measurement corridor.

For each segment, the left road edge is used as the OpenDRIVE reference line. The selected edge points are first fitted by a cubic curve in the mapping frame. The fitted curve is then converted into the OpenDRIVE ParamPoly3\footnote{ParamPoly3 is an OpenDRIVE geometry representation in which the local reference-line coordinates are expressed as cubic polynomial functions of a normalized curve parameter.} representation~\cite{ASAM_OpenDRIVE_190}, defined by the segment start position $(x_0, y_0)$, initial heading $h_0$, segment length $L$, and a normalized curve parameter $p \in [0,1]$:
\begin{align}
    u(p) &= a_U + b_U p + c_U p^2 + d_U p^3 \\
    v(p) &= a_V + b_V p + c_V p^2 + d_V p^3,
\end{align}
where $(u,v)$ denotes the local coordinate system of the reference line segment. The coefficient sets $(a_U,b_U,c_U,d_U)$ and $(a_V,b_V,c_V,d_V)$ are the fitted cubic ParamPoly3 coefficients that describe the variation of the reference line in the local $u$- and $v$-directions, respectively. The inertial coordinates of the segment are obtained by rotating and translating $(u,v)$ with $h_0$ and $(x_0,y_0)$. Together with the segment length $L$, these coefficients define the OpenDRIVE reference-line geometry used for map generation.

\subsubsection{Elevation Profile Reconstruction}
The mainline is reconstructed as a three-dimensional road representation rather than only a planar curve. After the horizontal reference line is fitted, height measurements from the vehicle pose trajectory and road edge point clouds are associated with sampled positions along the reference line. For a sampled reference point $C(s_i)=(x(s_i),y(s_i))$, nearby measurement points are selected in the horizontal plane:
\begin{equation}
    \mathcal{N}_i = \{(x_j,y_j,z_j) \mid \|(x_j,y_j)-C(s_i)\|_2 \le r\},
\end{equation}
where $(x_j,y_j,z_j)$ is a measurement point with height information and $r$ is the search radius. The elevation at $s_i$ is estimated using the median height of the neighboring points,
\begin{equation}
    \hat{z}(s_i)=\operatorname{median}_{(x_j,y_j,z_j)\in\mathcal{N}_i}(z_j),
\end{equation}
which reduces the influence of isolated noisy measurements from the point cloud.

The sampled elevations are then smoothed and fitted to the OpenDRIVE elevation representation~\cite{ASAM_OpenDRIVE_190}. For each reference line segment, the vertical profile is modeled as a cubic polynomial with respect to the local longitudinal coordinate $\Delta s$:
\begin{equation}
    z(\Delta s)=a_e+b_e\Delta s+c_e\Delta s^2+d_e\Delta s^3.
\end{equation}
Here, $a_e$ denotes the segment start elevation, $b_e$ denotes the initial grade, and $c_e$ and $d_e$ describe the change of grade along the segment. The fitted elevation coefficients are stored together with the horizontal reference line parameters so that the exported OpenDRIVE road contains both horizontal and vertical mainline geometry.

\subsubsection{Lane Sample Transformation}
Lane marking samples are initially represented in the ego vehicle coordinate frame of each LiDAR frame. To combine samples from different frames, each point $(x_{ego}, y_{ego})$ is transformed into the mapping frame using the synchronized vehicle pose $(x_0, y_0, \theta)$:
\begin{equation}
\begin{bmatrix} x_{map} \\ y_{map} \end{bmatrix} = 
\begin{bmatrix} \cos\theta & -\sin\theta \\ \sin\theta & \cos\theta \end{bmatrix} 
\begin{bmatrix} x_{ego} \\ y_{ego} \end{bmatrix} + 
\begin{bmatrix} x_0 \\ y_0 \end{bmatrix}.
\end{equation}
Here, $(x_0,y_0)$ denotes the vehicle position in the mapping frame and $\theta$ is the recorded yaw angle. This transformation aggregates sparse lane observations from individual frames into the same coordinate frame as the fitted reference line, enabling lane boundary estimation along each mainline segment.

\subsubsection{Lane Boundary Fitting}
The transformed lane samples are projected onto the fitted reference line using a local Frenet coordinate frame~\cite{Werling2010Frenet}, which is presented as road reference line coordinate system in~\autoref{fig:coordinate_frames}. For a lane sample point $P$, the longitudinal coordinate of road reference line system is obtained from the closest point on the reference line $C(s)$:
\begin{equation}
    s^* = \operatorname*{arg\,min}_{s} \| P - C(s) \|_2.
\end{equation}
The lateral offset is then computed as the signed distance from $P$ to $C(s^*)$ along the local normal direction. Points outside the longitudinal range of the current segment or far from the reference line are discarded to keep only lane samples within the reconstruction corridor.

The remaining lane samples are clustered according to their lateral offsets using K-Means clustering, which separates parallel lane boundaries in the Frenet frame. The number of clusters is selected automatically from candidate values up to six by maximizing the silhouette score~\cite{Rousseeuw1987Silhouettes}. The resulting clusters are ordered by their mean lateral offset to obtain a consistent lane boundary sequence. For each detected lane boundary, the lateral offset is modeled as a cubic function of the longitudinal coordinate of road reference line system:
\begin{equation}
    t_{lane}(s) = a_{lane} + b_{lane}s + c_{lane}s^2 + d_{lane}s^3,
\end{equation}
where $a_{lane}$, $b_{lane}$, $c_{lane}$, and $d_{lane}$ are estimated by least squares fitting. The fitted lane offset coefficients are stored together with the reference line parameters and used to construct the OpenDRIVE lane sections of the mainline.

\subsection{Ramp Reconstruction}

This section describes how unmeasured ramp segments are reconstructed from the preprocessed OSM road graph introduced in~\autoref{subsec:osm_data}. The reconstruction first identifies ramp candidates from the directed graph, then infers their lane attributes and topological type, and finally converts their OSM geometries into OpenDRIVE-compatible ramp representations.

\subsubsection{Ramp Candidate Identification}
As shown in~\autoref{fig:LiDAR_ramp_overview}, ramp segments outside the driven mainline are not directly available from the LiDAR point cloud. Ramp candidates are therefore identified from the local OSM road graph queried around the measured trajectory. In OSM, road classes with the \textit{\_link} suffix denote connector roads between major roads. Directed edges with such tags, including \textit{motorway\_link} and \textit{trunk\_link}, are therefore treated as initial ramp candidates.

Because not every nearby connector belongs to the measured highway segment, the semantic candidates are further checked against the measured trajectory in the local metric frame. Candidates whose geometries converge to the trajectory corridor are retained, and their identifiers, directed endpoint nodes, polyline geometries, and metadata are passed to subsequent attribute inference and geometry fitting, as summarized in Algorithm~\ref{alg:ramp_candidate}.

{\color{black}
\begin{algorithm}[ht]
\caption{Ramp Candidate Identification from OSM}
\label{alg:ramp_candidate}
\begin{algorithmic}[1]
\Require Preprocessed OSM road graph $G=(V,E)$ with edge highway tag $h_e$, measured trajectory $T_m$, distance threshold $d_r$
\Ensure Ramp candidate set $\mathcal{R}$
\State $\mathcal{R} \gets \emptyset$
\For{each directed edge $e \in E$}
    \If{\textit{\_link} $\notin$ $h_e$}
        \State \textbf{continue}
    \EndIf
    \State Project the edge geometry $g_e$ into the local metric frame
    \State $d_e \gets \min_{p \in g_e,\, q \in T_m}\|p-q\|_2$
    \If{$d_e \le d_r$}
        \State Preserve the edge ID, endpoint nodes, geometry, and metadata
        \State $\mathcal{R} \gets \mathcal{R} \cup \{e\}$
    \EndIf
\EndFor
\State \Return $\mathcal{R}$
\end{algorithmic}
\end{algorithm}
}

\subsubsection{Lane Count Inference}
Lane count is required to generate ramp boundaries and OpenDRIVE lane sections. The driving lane count of each relevant OSM edge is estimated using the fields in~\autoref{tab:osm_fields}. Because these tags are not always complete or consistently populated, a priority rule is used, as summarized in Algorithm~\ref{alg:lane_inference}.
\begin{table}[ht]
\centering
\caption{OSM fields used for lane count inference.}
\label{tab:osm_fields}
\begin{tabular}{l >{\linespread{0.85}\selectfont}m{0.65\linewidth}} \toprule[2pt] 
\textbf{OSM Field} & \textbf{Description} \\ \midrule[1pt]
\textit{lanes:forward} & Number of lanes in the travel direction \\
\textit{lanes} & Total number of lanes on the road segment \\
\textit{turn:lanes} & Turn designation for each lane, separated by \textit{|} \\
\textit{highway} & Road classification (e.g. motorway, trunk, primary) \\ \bottomrule[2pt]
\end{tabular}
\end{table}

{\color{black}
\begin{algorithm}[ht]
\caption{OSM Edge Lane Count Inference}
\label{alg:lane_inference}
\begin{algorithmic}[1]
\Require OSM edge tag dictionary $T$
\Ensure Estimated edge lane count $N_e$

\State $v_f \gets \text{ParseInteger}(T[\textit{'lanes:forward'}])$
\State $v_l \gets \text{ParseInteger}(T[\textit{'lanes'}])$
\If{$v_f$ is valid}
    \State $N_e \gets v_f$
\ElsIf{$v_l$ is valid}
    \State $N_e \gets v_l$
\ElsIf{\textit{'turn:lanes'} $\in T$}
    \State $S \gets T[\textit{'turn:lanes'}]$
    \State $N_e \gets \text{Count}(S, \text{'|'}) + 1$ 
\Else
    \State $H \gets T[\textit{'highway'}]$ 
    \If{\textit{'motorway'} $\in H$}
        \State $N_e \gets 3$
    \ElsIf{\textit{'trunk'} $\in H$ \textbf{or} \textit{'primary'} $\in H$}
        \State $N_e \gets 2$
    \Else
        \State $N_e \gets 1$
    \EndIf
\EndIf

\State \Return $N_e$
\end{algorithmic}
\end{algorithm}
}

The ramp-level count is then assigned from the local mainline lane-count change near the ramp event. An increase indicates lanes entering through a merge ramp, whereas a decrease indicates lanes leaving through an exit ramp. At least one lane is assigned to each ramp, and the resulting count determines the number of ramp boundaries used in geometry fitting.

\subsubsection{Merge and Exit Classification}
Each confirmed ramp candidate is classified as either a \textit{Merge} or an \textit{Exit} using the directed connectivity of the OSM graph. Let $E_m=(u_m,v_m)$ denote the connected mainline edge and $E_r=(u_r,v_r)$ denote the ramp edge. The ramp type is determined by
\begin{equation}
T_{ramp} =
\begin{cases}
\text{Merge}, & v_r \in \{u_m,v_m\},\\
\text{Exit}, & \text{otherwise}.
\end{cases}
\end{equation}
This rule reflects whether the directed ramp edge terminates at the mainline or leaves it. The inferred type is used in the following integration stage to determine the connection direction and mainline-ramp topology.

\subsubsection{Ramp Geometry Reconstruction}
For each confirmed ramp, the OSM polyline is used as the initial ramp centerline. If an OSM edge does not contain an explicit polyline geometry, the source and target nodes define the initial line. Because OSM geometries may be sparse and locally angular, the centerline is uniformly resampled and smoothed before parametric fitting. In the implementation, a \SI{0.5}{\meter} sampling interval and a Savitzky-Golay filter are used to obtain a dense and smooth ramp centerline.

The inferred lane count is then used to generate ramp lane boundaries from the smoothed centerline. Let $C_r(s_i)$ be the ramp centerline point at longitudinal position $s_i$, $\mathbf{n}(s_i)$ be its local unit normal vector, $N_r$ be the ramp lane count, $w_r$ be the lane width, and $B_k(s_i)$ be the $k$th boundary point. The $N_r+1$ lane boundaries are generated as
\begin{equation}
    B_k(s_i)=C_r(s_i)+\left(\frac{N_r}{2}-k\right)w_r\mathbf{n}(s_i), \quad k=0,\ldots,N_r.
\end{equation}
Here, $k=0$ corresponds to the leftmost boundary and $k=N_r$ to the rightmost boundary. Following the mainline representation, the leftmost boundary is used as the ramp reference line, and the remaining boundaries are represented as lateral offsets in the local Frenet frame. The ramp reference line and lane offsets are fitted by polynomial functions, yielding OpenDRIVE-compatible ramp road parameters for the subsequent integration and export stages.

\subsection{Mainline-Ramp Integration}

After the LiDAR-derived mainline and OSM-derived ramps are reconstructed separately, they must be organized into a unified road network. This stage determines how each ramp connects to the adjacent mainline segment according to its inferred ramp type and directed OSM connectivity, producing the road-level topology required for OpenDRIVE export.

\subsubsection{Connection Endpoint Determination}
For each reconstructed ramp, the connection direction is determined by the ramp type identified in Section~III-C. An exit ramp is treated as a diverging connection from the mainline to the ramp, whereas a merge ramp is treated as a converging connection from the ramp to the mainline. Let $R_m$ denote the adjacent mainline road and $R_r$ denote the reconstructed ramp road. The road-level connection is defined as
\begin{equation}
    \mathcal{C}(R_r)=
    \begin{cases}
    R_m \rightarrow R_r, & T_{ramp}=\text{Exit},\\
    R_r \rightarrow R_m, & T_{ramp}=\text{Merge}.
    \end{cases}
\end{equation}
The corresponding connection endpoint is selected from the start or end of the ramp centerline according to this direction. This produces an explicit mainline-ramp relation while preserving the directed topology inherited from the OSM graph.

\subsubsection{Road Topology Construction}
After the connection direction and endpoint are determined, the mainline segments and ramp roads are assembled into a directed road graph. Each road is represented as a node, and each valid driving connection is represented as a directed edge. The topology rules used for ramp integration are summarized in~\autoref{tab:ramp_topology}.

\begin{table}[ht]
\centering
\caption{Topology rules for mainline-ramp integration.}
\label{tab:ramp_topology}
\begin{tabular}{lccc} \toprule[2pt]
\textbf{Ramp Type} & \textbf{Connection} & \textbf{Ramp Endpoint} & \textbf{Topology Role} \\ \midrule[1pt]
Exit & $R_m \rightarrow R_r$ & Start & Diverging \\
Merge & $R_r \rightarrow R_m$ & End & Converging \\ \bottomrule[2pt]
\end{tabular}
\end{table}

This topology separates geometric reconstruction from network connectivity. The geometry of each road remains defined by its fitted reference line and lane offsets, whereas the graph edges describe how vehicles can move between roads. The resulting directed road graph is used in the subsequent OpenDRIVE export stage to construct road predecessors, successors, and junction connections.

\subsection{OpenDRIVE Export}

The reconstructed map is exported as an ASAM OpenDRIVE file, which provides a standardized XML representation for road geometry, lane structure, elevation, and road connectivity~\cite{ASAM_OpenDRIVE_190}. The mapping between workflow-derived geometry and OpenDRIVE elements is summarized in~\autoref{tab:opendrive_mapping}.

\begin{table}[ht]
\centering
\caption{Mapping from workflow-derived geometry to OpenDRIVE representations.}
\label{tab:opendrive_mapping}
\begin{tabular}{ll} \toprule[2pt]
\textbf{Workflow-derived Geometry} & \textbf{OpenDRIVE Element} \\ \midrule[1pt]
Mainline geometry (III-B) & \textit{road}, \textit{planView} \\
Mainline elevation (III-B) & \textit{elevationProfile} \\
Mainline lanes (III-B) & \textit{laneSection} \\
Ramp geometry (III-C) & \textit{road}, \textit{planView} \\
Ramp lanes (III-C) & \textit{laneSection} \\
Integrated topology (III-D) & \textit{link}, \textit{junction} \\
Coordinate reference (III-A) & \textit{geoReference} \\ \bottomrule[2pt]
\end{tabular}
\end{table}

\begin{figure}[ht]
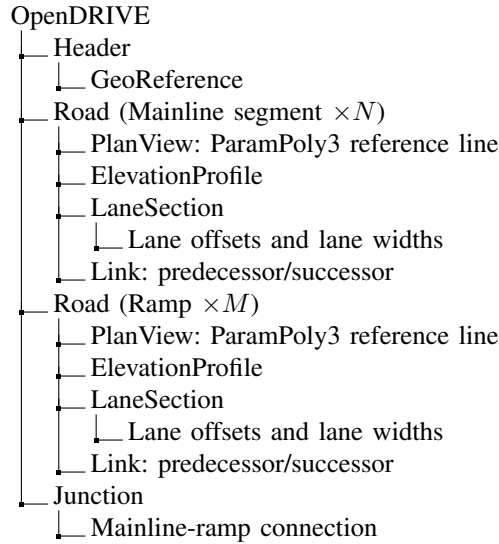

    \centering
    \begin{minipage}{\columnwidth}
        \renewcommand*\DTstyle{\rmfamily}
        \dirtree{%
        .1 OpenDRIVE.
        .2 Header.
        .3 GeoReference.
        .2 Road (Mainline segment $\times N$).
        .3 PlanView: ParamPoly3 reference line.
        .3 ElevationProfile.
        .3 LaneSection.
        .4 Lane offsets and lane widths.
        .3 Link: predecessor/successor.
        .2 Road (Ramp $\times M$).
        .3 PlanView: ParamPoly3 reference line.
        .3 ElevationProfile.
        .3 LaneSection.
        .4 Lane offsets and lane widths.
        .3 Link: predecessor/successor.
        .2 Junction.
        .3 Mainline-ramp connection.
        }
    \end{minipage}
    \caption{Hierarchical structure of the generated OpenDRIVE digital twin.}
    \label{fig:opendrive_structure}
\end{figure}

As illustrated in~\autoref{fig:opendrive_structure}, these elements are organized into a complete OpenDRIVE hierarchy. The components are assembled using the \textit{scenariogeneration} Python library~\cite{scenariogeneration_github} and exported as a simulation-ready road network.

\section{RESULT}

\subsection{Qualitative Evaluation}
\begin{figure*}[!t]
    \centering
    \begin{tikzpicture}[
        panel/.style={
            draw=black!35,
            line width=0.3pt,
            minimum width=0.315\textwidth,
            minimum height=0.19\textwidth,
            inner sep=0pt,
            outer sep=0pt
        },
        sublabel/.style={
            anchor=north west,
            fill=white,
            fill opacity=0.85,
            text opacity=1,
            inner sep=2pt,
            font=\scriptsize
        }
    ]
        \node[panel] (a) at (0,0)
        {\includegraphics[width=0.315\textwidth,height=0.19\textwidth,keepaspectratio]{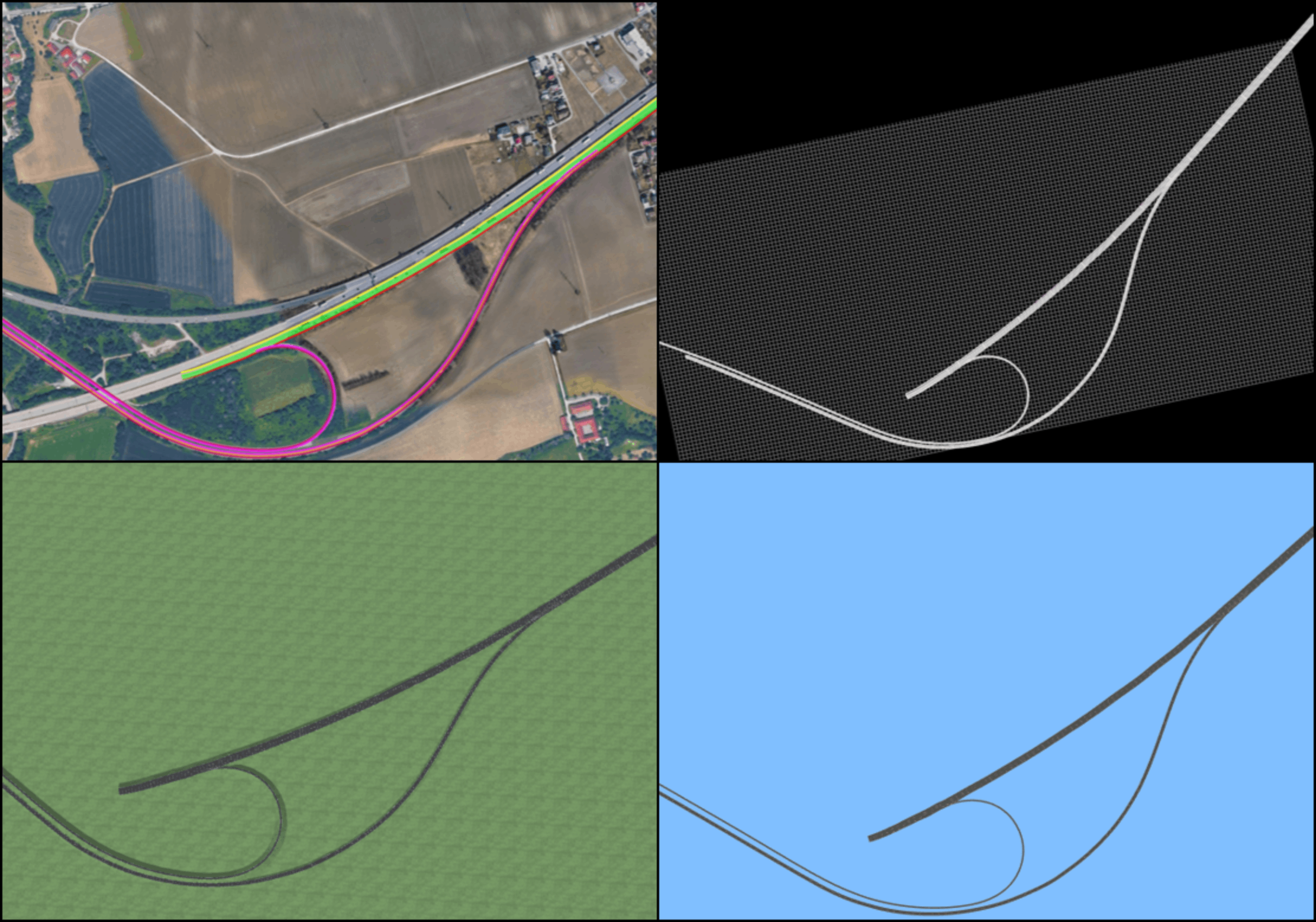}};
        \node[panel] (b) at (0.34\textwidth,0)
        {\includegraphics[width=0.315\textwidth,height=0.19\textwidth,keepaspectratio]{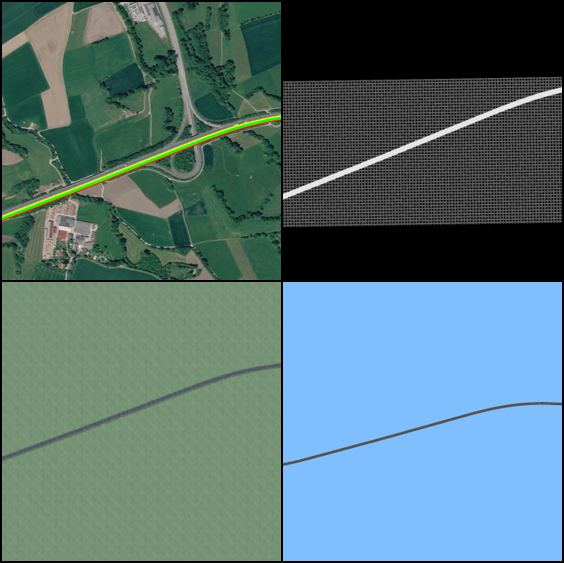}};
        \node[panel] (c) at (0.68\textwidth,0)
        {\includegraphics[width=0.315\textwidth,height=0.19\textwidth,keepaspectratio]{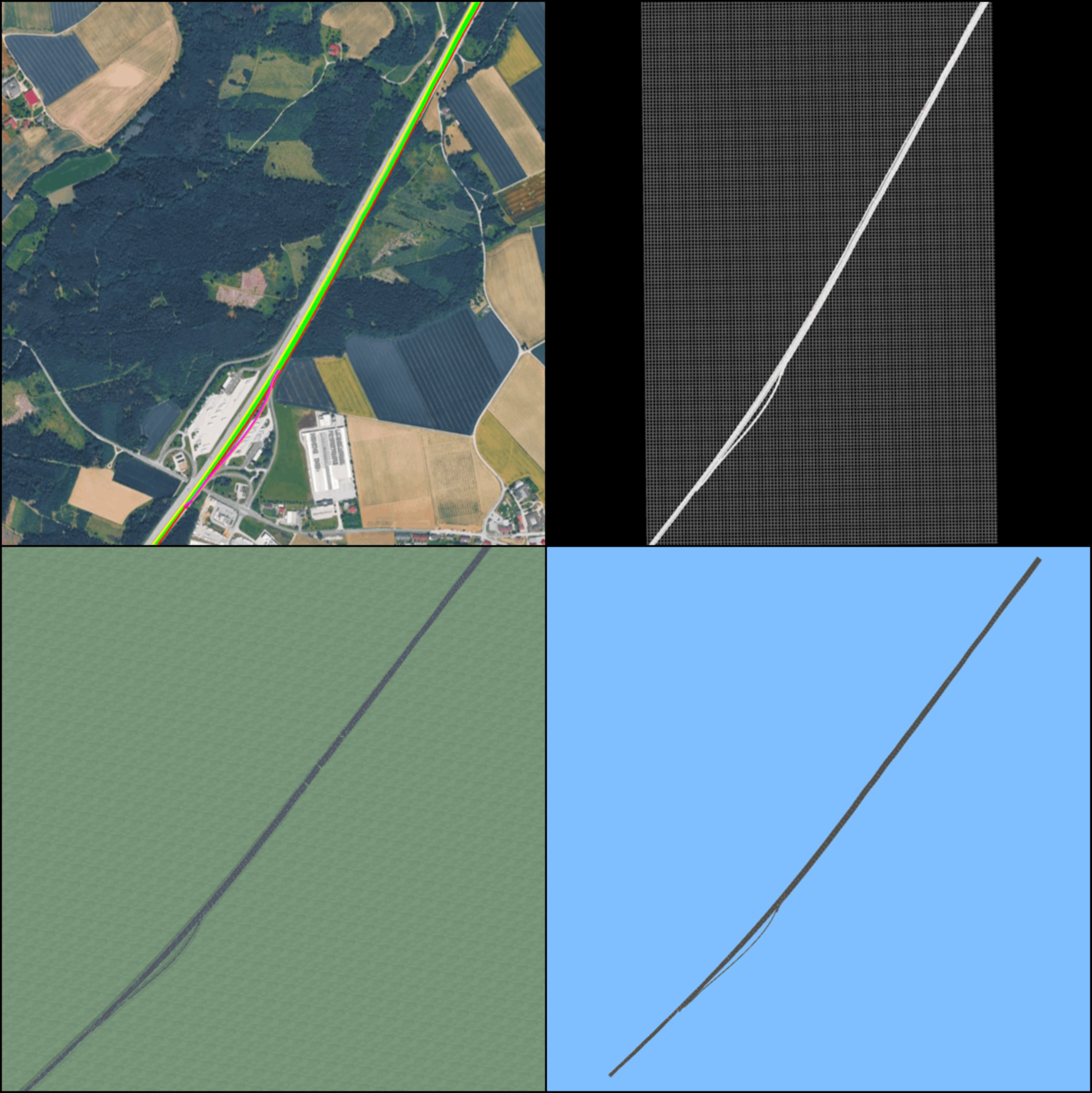}};
        \node[panel] (d) at (0,-0.22\textwidth)
        {\includegraphics[width=0.315\textwidth,height=0.19\textwidth,keepaspectratio]{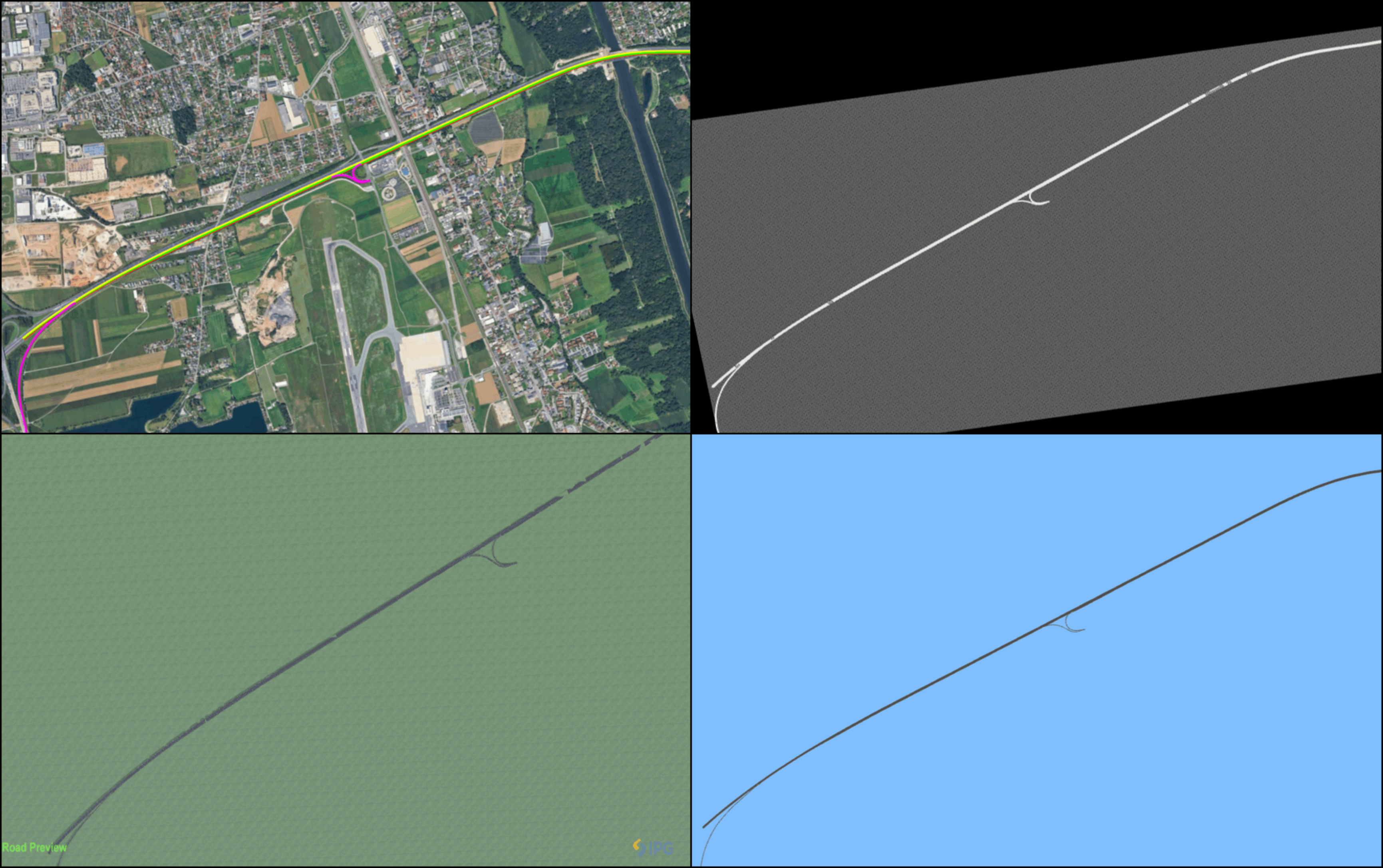}};
        \node[panel] (e) at (0.34\textwidth,-0.22\textwidth)
        {\includegraphics[width=0.315\textwidth,height=0.19\textwidth,keepaspectratio]{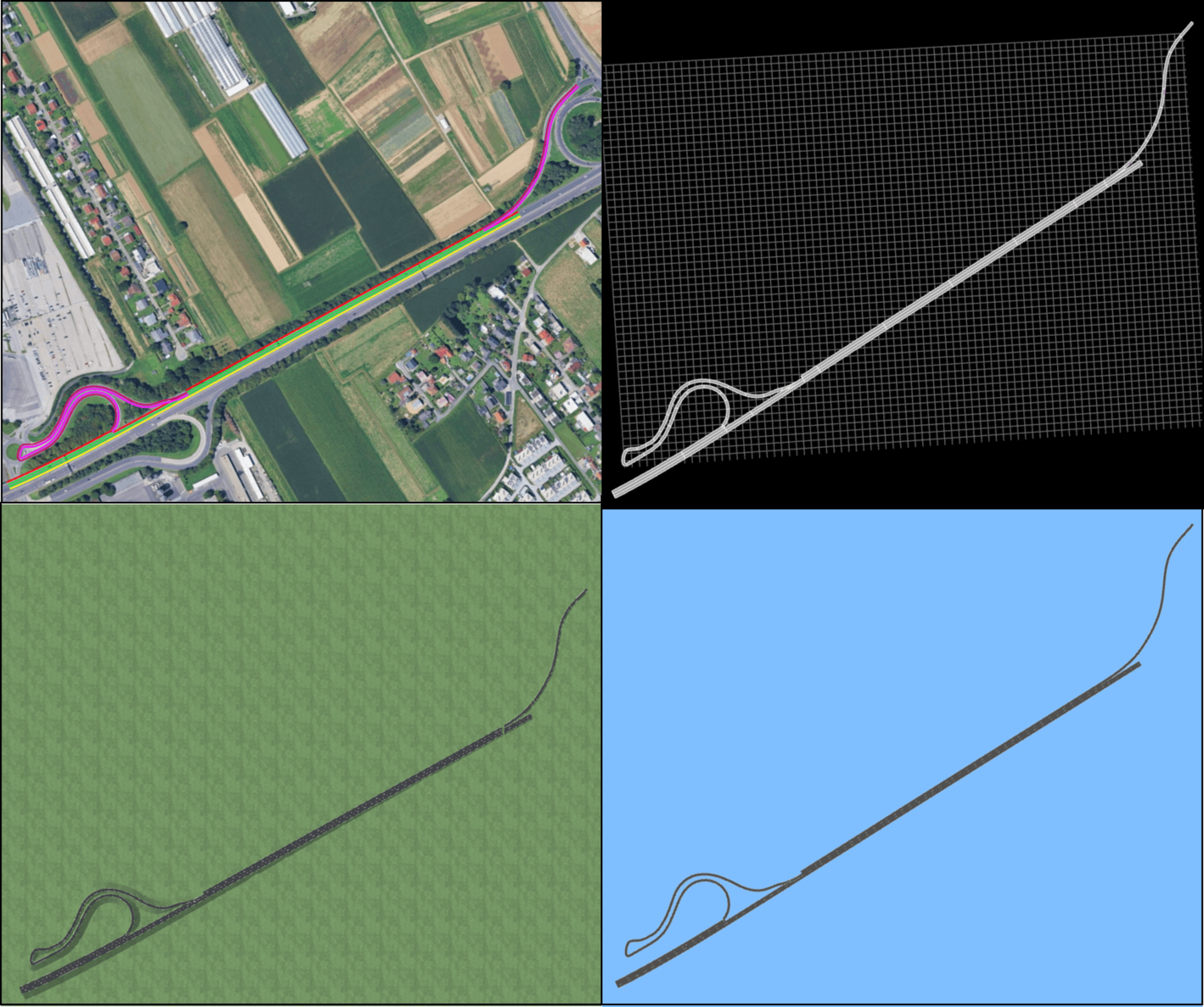}};
        \node[panel] (f) at (0.68\textwidth,-0.22\textwidth)
        {\includegraphics[width=0.315\textwidth,height=0.19\textwidth,keepaspectratio]{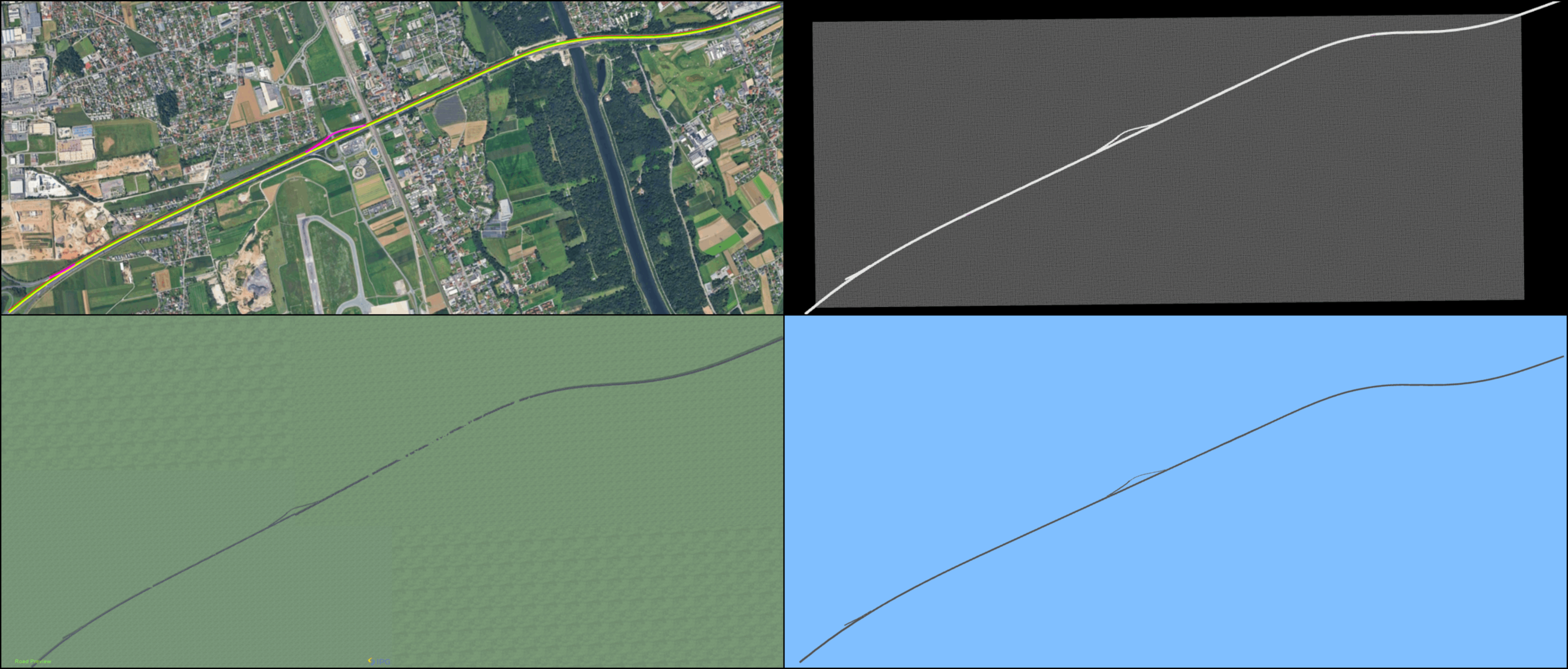}};

        \node[sublabel] at (a.north west) {(a)};
        \node[sublabel] at (b.north west) {(b)};
        \node[sublabel] at (c.north west) {(c)};
        \node[sublabel] at (d.north west) {(d)};
        \node[sublabel] at (e.north west) {(e)};
        \node[sublabel] at (f.north west) {(f)};
    \end{tikzpicture}
    \caption{Representative qualitative visualization of six generated OpenDRIVE digital twins in different highway environments. Panels (a)--(f) correspond to six reconstructed cases. In each panel, the top-left view shows the road geometry extracted from the generated XODR file and overlaid on Google Earth imagery for geographic inspection; the top-right view shows the rendering in OpenDRIVE Viewer; the bottom-left view shows the visualization in IPG CarMaker; and the bottom-right view shows the visualization in Esmini.}
    \label{fig:qualitative_result}
\end{figure*}
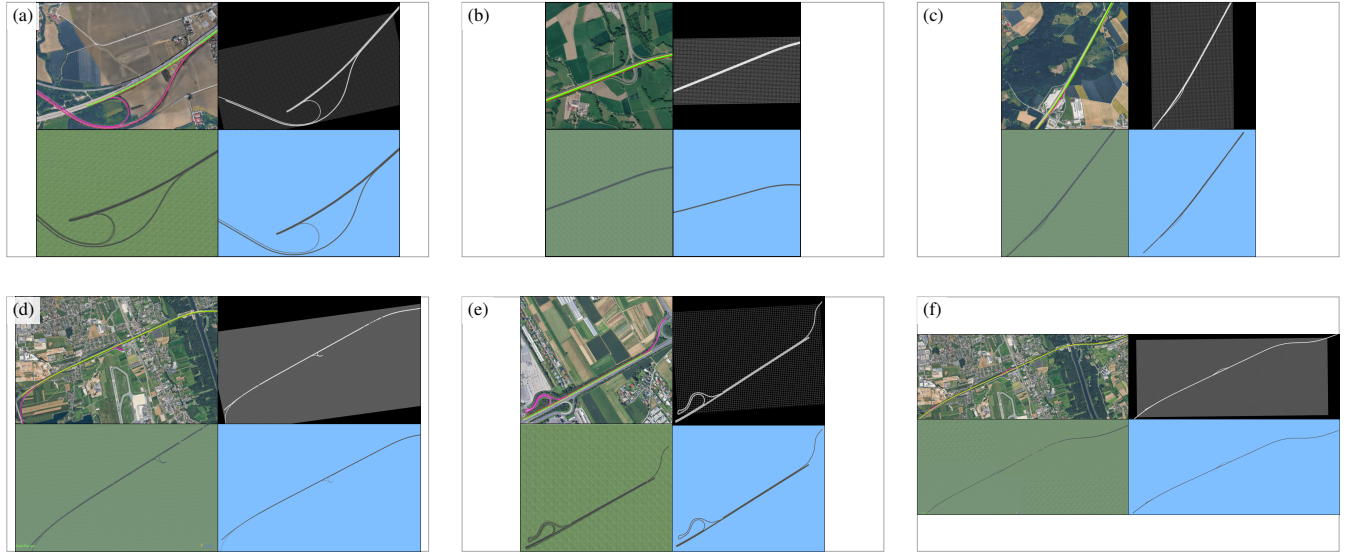

To qualitatively evaluate the reconstructed digital twin, the generated ASAM OpenDRIVE road model was imported and visualized in Google Earth\footnote{The OpenDRIVE road geometry was extracted and converted to Keyhole Markup Language (KML) format, a standard geospatial data format supported by Google Earth, which was subsequently imported for spatial visualization of the reconstructed road network.}, OpenDRIVE Viewer~\cite{ODRViewer},~IPG CarMaker~\cite{IPG_CarMaker} (a high-fidelity vehicle dynamics testing platform), and Esmini~\cite{esmini_github} (an open-source OpenDRIVE simulator), as shown in~\autoref{fig:qualitative_result}. In the Google Earth overlays, the reconstructed geometry follows the corresponding highway corridors, with green segments denoting the LiDAR-derived mainline and red segments denoting OSM-derived ramps. The spatially consistent placement of the ramps relative to the mainline indicates that they are transformed into the same coordinate frame and connected at plausible merge or exit locations. The same OpenDRIVE files are also rendered in OpenDRIVE Viewer, IPG CarMaker, and Esmini without manual modification, showing that the generated geometry, lane sections, and road links/junctions can be interpreted by common OpenDRIVE-based tools. 

\subsection{Quantitative Geometric Evaluation}
The quantitative geometric evaluation assesses the lateral consistency between the generated OpenDRIVE boundaries and reference measurements extracted from the mobile mapping data. These reference measurements include LiDAR-derived road edge points and lane marking samples. The evaluation was performed in the road reference line coordinate system. For each measurement point, the lateral error was computed as the distance between its lateral coordinate and the corresponding OpenDRIVE boundary at the same longitudinal position:
\begin{equation}
    e_i=\left|t_{m,i}-t_b(s_{m,i})\right|,
\end{equation}
where $t_{m,i}$ is the lateral coordinate of the $i$th measurement point, $s_{m,i}$ is its longitudinal coordinate, and $t_b(s_{m,i})$ is the lateral coordinate of the sampled OpenDRIVE boundary at the same longitudinal position. Road edges and lane boundaries were evaluated separately because they represent different levels of map geometry: the outer drivable-area envelope and the lane-level structure inside it.

\begin{table}[ht]
    \centering
    \caption{Summary of Frenet lateral accuracy over six real-world highway sequences.}
    \label{tab:quantitative_accuracy}
    \footnotesize
    \begin{tabular}{lccc} \toprule[2pt]
        \makecell{\textbf{Evaluation}\\\textbf{Target}} &
        \makecell{\textbf{Mean}\\\textbf{RMSE (m)}} &
        \makecell{\textbf{Median}\\\textbf{RMSE (m)}} &
        \makecell{\textbf{RMSE}\\\textbf{Range (m)}} \\
        \midrule[1pt]
        Road edge & 0.596 & 0.476 & 0.099--1.235 \\
        Lane boundary & 0.883 & 0.996 & 0.262--1.114 \\
        Overall & 0.740 & 0.717 & 0.356--1.115 \\
        \bottomrule[2pt]
    \end{tabular}
\end{table}

\autoref{tab:quantitative_accuracy} summarizes the evaluation results over six real-world highway sequences. The generated maps achieved an overall mean RMSE of 0.740~m, with sequence-level overall RMSE values ranging from 0.356~m to 1.115~m. The road-edge accuracy was slightly higher than the lane-boundary accuracy, with mean RMSE values of 0.596~m and 0.883~m, respectively. This difference is expected because road edges are directly constrained by LiDAR-derived boundary observations, whereas lane boundaries are inferred from sparse lane samples and clustering-based lateral offsets. These results indicate sub-meter to near-meter lateral accuracy for simulation-oriented OpenDRIVE digital twin generation.

\section{CONCLUSION}
This paper presented an automated workflow for generating simulation-oriented OpenDRIVE digital twins in highway scenarios by combining mobile mapping LiDAR point clouds and OSM road graphs. The workflow reconstructs the measured mainline, including the reference line, elevation profile, and lane boundaries, from LiDAR-derived measurements, while OSM provides complementary geometric and topological information for ramp segments outside the measurement corridor. Ramp candidates, lane attributes, and merge or exit types are inferred from the OSM road graph, and the reconstructed mainline and ramps are integrated through directed topology construction before being exported as a georeferenced ASAM OpenDRIVE map.

The experimental results demonstrate the effectiveness of the proposed workflow for generating usable simulation maps from highway measurements. Qualitative visualization in Google Earth, OpenDRIVE Viewer, IPG CarMaker, and Esmini confirms that the generated road geometry and ramp topology can be consistently inspected across different tools. Quantitatively, the generated maps achieved an overall mean lateral RMSE of 0.740~m over six real-world highway sequences, demonstrating sub-meter geometric consistency for simulation-oriented digital twin generation. The proposed approach still relies on the completeness and geometric quality of OSM data and on the density of LiDAR-derived road evidence. Future work will extend the evaluation to more complex road topologies.



\bibliographystyle{IEEEtran}
\bibliography{ref}

@ARTICLE{10417138,
  author={Zhao, Yongqi and Burghardt, Tomasz E. and Li, Hexuan and Rosenberger, Philipp and Babić, Dario and Babić, Darko and Wiesinger, Friedrich and Helmreich, Bernhard and Eichberger, Arno},
  journal={IEEE Sensors Journal}, 
  title={Enhancing {LiDAR} Reliability Through Utilization of Premium Road Marking Materials}, 
  year={2024},
  volume={24},
  number={6},
  pages={8015-8025},
  doi={10.1109/JSEN.2024.3359754}}

@ARTICLE{11361285,
  author={Zhao, Yongqi and Zhou, Ji and Bi, Dong and Mihalj, Tomislav and Hu, Jia and Eichberger, Arno},
  journal={IEEE Transactions on Intelligent Transportation Systems}, 
  title={A Survey on the Application of Large Language Models in Scenario-Based Testing of Automated Driving Systems}, 
  year={2026},
  volume={27},
  number={5},
  pages={5001-5023},
  doi={10.1109/TITS.2026.3651004}}

@article{KALRA2016182,
title = {Driving to safety: How many miles of driving would it take to demonstrate autonomous vehicle reliability?},
journal = {Transportation Research Part A: Policy and Practice},
volume = {94},
pages = {182-193},
year = {2016},
issn = {0965-8564},
doi = {https://doi.org/10.1016/j.tra.2016.09.010},
author = {Nidhi Kalra and Susan M. Paddock}
}

@article{boeing2025modeling,
  title={Modeling and analyzing urban networks and amenities with {OSMnx}},
  author={Boeing, Geoff},
  journal={Geographical Analysis},
  volume={57},
  number={4},
  pages={567--577},
  year={2025},
  publisher={Wiley Online Library}
}

@misc{ASAM_OpenDRIVE_190,
  author       = {{ASAM e.V.}},
  title        = {{ASAM OpenDRIVE BS 1.9.0 Specification}},
  year         = {2026},
  howpublished = {Online},
  url          = {https://publications.pages.asam.net/standards/ASAM_OpenDRIVE/ASAM_OpenDRIVE_Specification/latest/specification/index.html},
  note         = {Accessed: 2026-05-28}
}

@inproceedings{Werling2010Frenet,
  author    = {Werling, Moritz and Ziegler, Julius and Kammel, S{\"o}ren and Thrun, Sebastian},
  title     = {Optimal Trajectory Generation for Dynamic Street Scenarios in a {Frenet} Frame},
  booktitle = {Proceedings of the 2010 IEEE International Conference on Robotics and Automation},
  year      = {2010},
  pages     = {987--993},
  doi       = {10.1109/ROBOT.2010.5509799}
}

@article{Rousseeuw1987Silhouettes,
  author  = {Rousseeuw, Peter J.},
  title   = {Silhouettes: A Graphical Aid to the Interpretation and Validation of Cluster Analysis},
  journal = {Journal of Computational and Applied Mathematics},
  volume  = {20},
  pages   = {53--65},
  year    = {1987},
  doi     = {10.1016/0377-0427(87)90125-7}
}

@misc{scenariogeneration_github,
  author       = {Andersson, Mikael and Natale, Irene and Tingberg, Andreas and Kaths, Jakob and Kelidari, Esmat},
  title        = {{scenariogeneration}: Python Library to Generate Linked {OpenDRIVE} and {OpenSCENARIO} Files},
  year         = {2026},
  howpublished = {Online},
  url          = {https://github.com/pyoscx/scenariogeneration},
  note         = {Accessed: 2026-05-30}
}

@misc{esmini_github,
  author       = {{esmini contributors}},
  title        = {{Environment Simulator Minimalistic (esmini)}: A Basic {OpenSCENARIO} Player},
  year         = {2026},
  howpublished = {Online},
  url          = {https://github.com/esmini/esmini},
  note         = {Accessed: 2026-05-30}
}

@misc{IPG_CarMaker,
  author       = {{IPG Automotive GmbH}},
  title        = {{CarMaker}: The Simulation Solution for Virtual Test Driving},
  year         = {2025},
  howpublished = {Online},
  url          = {https://www.ipg-automotive.com/solutions/product-portfolio/carmaker},
  note         = {Accessed: 2026-05-30}
}

@misc{ODRViewer,
  author       = {{odrviewer.io}},
  title        = {{odrviewer.io}: Online {OpenDRIVE} File Viewer},
  year         = {2026},
  howpublished = {Online},
  url          = {https://odrviewer.io/},
  note         = {Accessed: 2026-05-30}
}

@ARTICLE{10242366,
  author={Hu, Xuemin and Li, Shen and Huang, Tingyu and Tang, Bo and Huai, Rouxing and Chen, Long},
  journal={IEEE Transactions on Intelligent Vehicles}, 
  title={How Simulation Helps Autonomous Driving: A Survey of Sim2real, Digital Twins, and Parallel Intelligence}, 
  year={2024},
  volume={9},
  number={1},
  pages={593-612},
  doi={10.1109/TIV.2023.3312777}}

@article{barsi2021building,
  title={Building opendrive model from mobile mapping data},
  author={Barsi, M and Barsi, A},
  journal={The International Archives of the Photogrammetry, Remote Sensing and Spatial Information Sciences},
  volume={43},
  pages={9--14},
  year={2021},
  publisher={Copernicus GmbH}
}

@article{Elghazaly2023HDMaps,
  author  = {Elghazaly, Gamal and Frank, Raphael and Harvey, Scott and Safko, Stefan},
  title   = {High-Definition Maps: Comprehensive Survey, Challenges, and Future Perspectives},
  journal = {IEEE Open Journal of Intelligent Transportation Systems},
  volume  = {4},
  pages   = {527--550},
  year    = {2023},
  doi     = {10.1109/OJITS.2023.3295502}
}

@article{Liu2020HDMap,
  author  = {Liu, Rong and Wang, Jinling and Zhang, Bingqi},
  title   = {High Definition Map for Automated Driving: Overview and Analysis},
  journal = {The Journal of Navigation},
  volume  = {73},
  number  = {2},
  pages   = {324--341},
  year    = {2020},
  doi     = {10.1017/S0373463319000638}
}

@article{Chiang2022OpenDRIVE,
  author  = {Chiang, Kai-Wei and Pai, Hao-Yu and Zeng, Jhih-Cing and Tsai, Meng-Lun and El-Sheimy, Naser},
  title   = {Automated Modeling of Road Networks for High-Definition Maps in {OpenDRIVE} Format Using Mobile Mapping Measurements},
  journal = {Geomatics},
  volume  = {2},
  number  = {2},
  pages   = {221--235},
  year    = {2022},
  doi     = {10.3390/geomatics2020013}
}

@article{Wang2023LiDARRoadModeling,
  author  = {Wang, Yuchen and Wang, Weicheng and Liu, Jinzhou and Chen, Tianheng and Wang, Shuyi and Yu, Bin and Qin, Xiaochun},
  title   = {Framework for Geometric Information Extraction and Digital Modeling from {LiDAR} Data of Road Scenarios},
  journal = {Remote Sensing},
  volume  = {15},
  number  = {3},
  pages   = {576},
  year    = {2023},
  doi     = {10.3390/rs15030576}
}

@article{Somogyi2025AIHDMap,
  author  = {Somogyi, Arpad Jozsef and Baranyai, Daniel and Dowajy, Mohammad and Lovas, Tamas and Szalay, Zsolt and Tettamanti, Tamas},
  title   = {Artificial Intelligence Based High Definition Map Generation From Mobile Mapping Data},
  journal = {IEEE Access},
  volume  = {13},
  pages   = {121838--121848},
  year    = {2025},
  doi     = {10.1109/ACCESS.2025.3587592}
}

\end{document}